\documentclass[letterpaper]{article} 
\usepackage{aaai23}  
\usepackage{times}  
\usepackage{helvet}  
\usepackage{courier}  
\usepackage[hyphens]{url}  
\usepackage{graphicx} 
\usepackage{natbib}  
\usepackage{caption} 
\usepackage{algorithm}
\usepackage{algorithmic}
\usepackage{amsmath}
\usepackage{multirow}
\usepackage{setspace}

\usepackage{newfloat}
\usepackage{listings}
\DeclareCaptionStyle{ruled}{labelfont=normalfont,labelsep=colon,strut=off} 
\floatstyle{ruled}
\newfloat{listing}{tb}{lst}{}
\floatname{listing}{Listing}
\title{Learn More for Food Recognition via Progressive Self-Distillation}
\author {
    Yaohui Zhu\textsuperscript{\rm 1}\equalcontrib,
    Linhu Liu\textsuperscript{\rm 2}\equalcontrib, 
    Jiang Tian\textsuperscript{\rm 2}
}
\affiliations {
    \textsuperscript{\rm 1} School of Artifical Intelligance, Beijing Normal University,
Beijing 10875, China\\
    \textsuperscript{\rm 2} AI Lab, Lenovo Research, Beijing, China\\
    yaohui.zhu@bnu.edu.cn, \{liulh7, tianjiang1\}@lenovo.com
}

\usepackage{bibentry}

\begin{document}

\maketitle

\begin{abstract}
Food recognition has a wide range of applications, such as health-aware
recommendation and self-service restaurants. Most previous methods
of food recognition firstly locate informative regions in some weakly-supervised
manners and then aggregate their features. However, location errors
of informative regions limit the effectiveness of these methods to
some extent. Instead of locating multiple regions, we propose a Progressive
Self-Distillation (PSD) method, which progressively enhances the ability
of network to mine more details for food recognition. The training
of PSD simultaneously contains multiple self-distillations, in which
a teacher network and a student network share the same embedding network.
Since the student network receives a modified image from its teacher
network by masking some informative regions, the teacher network outputs
stronger semantic representations than the student network. Guided
by such teacher network with stronger semantics, the student network
is encouraged to mine more useful regions from the modified image
by enhancing its own ability. The ability of the teacher network is
also enhanced with the shared embedding network. By using progressive
training, the teacher network incrementally improves its ability to
mine more discriminative regions. In inference phase, only the teacher
network is used without the help of the student network. Extensive
experiments on three datasets demonstrate the effectiveness of our
proposed method and state-of-the-art performance.
\end{abstract}

\section{Introduction}

Food is a necessity of human life and the foundation of human experience.
As a basic research in food field, food recognition has a wide range
of applications such as visual food choice \citep{chen2020zero},
health-aware recommendation \citep{Nag-HML-ICMR2017} and self-service
restaurants \citep{Aguilar2018Grab}. 
\begin{figure}
\noindent \begin{centering}
\includegraphics[scale=0.63]{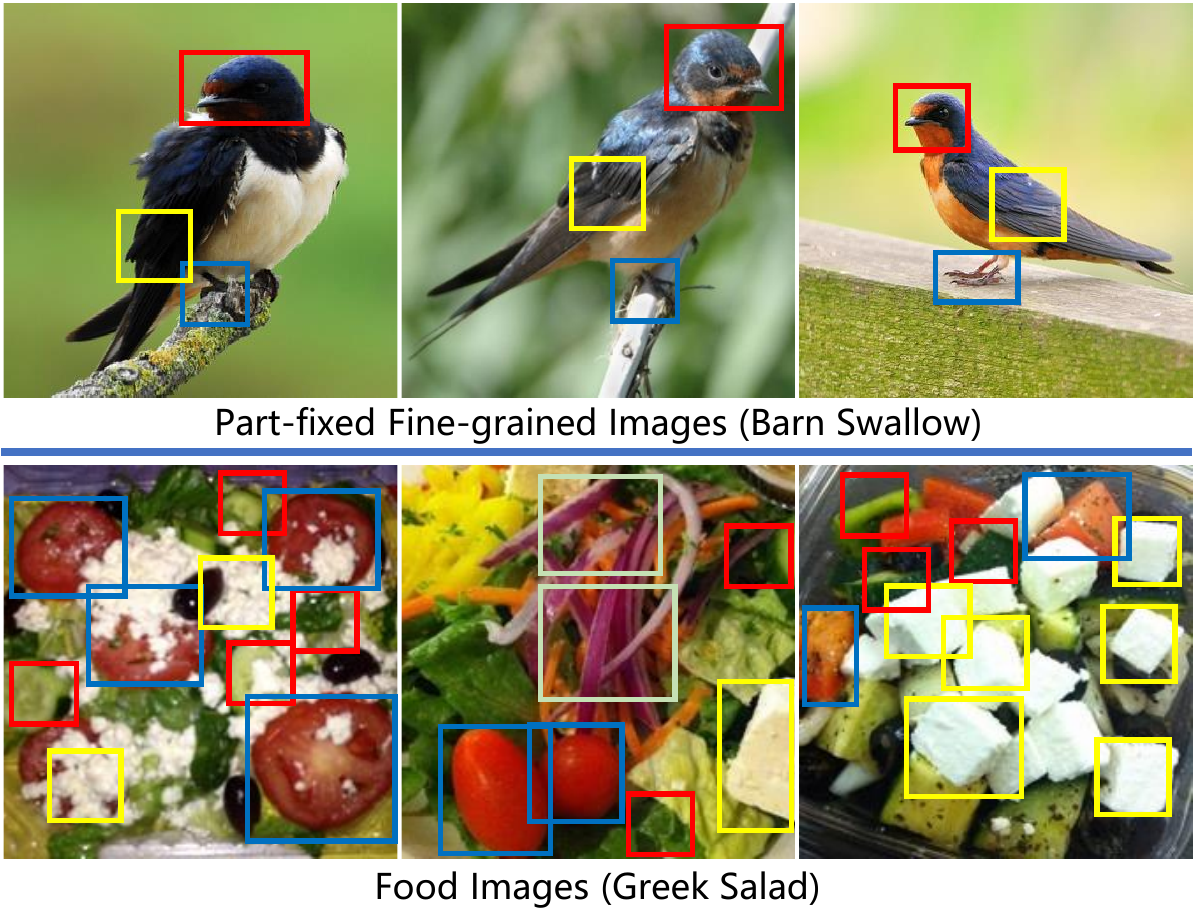}
\par\end{centering}
\caption{\label{samples_intro}Some fine-grained bird images and food images.
There may be three discriminative regions in `Barn Swallow' (see bounding
boxes), while `Greek Salad' contains lots of similar regions, each
of which is possibly discriminative. Therefore, we think that food
images have more informative regions than part-fixed fine-grained
images.}
\end{figure}

Food recognition belongs to fine-grained recognition, which refers
to the task of distinguishing subordinate categories, such as birds,
dogs and cars. Most previous methods of food recognition \citep{Martinel-WSR-WACV2018,Qiu-MDFR-BMVC2019,min2020isia,Min-IGCMAN-ACMMM2019,Wang2022}
mainly follow the key solution of fine-grained recognition, which
firstly locates informative regions in some weakly-supervised manners,
and then aggregates features of these regions. Although these methods
achieve promising performance, they possibly suffer from location
errors of those informative regions, leading to the limited effectiveness
to some extent. 

In comparison with part-fixed fine-grained images such as birds, food
images include a variety of similar ingredients stacked together.
As shown in Figure \ref{samples_intro}, there are some  semantic
parts (e.g., head and claw) in `Barn Swallow', while lots of sliced
material (e.g., cucumbers and cheese), fruit and vegetable are stacked
together in `Greek Salad'. An intuitive hypothesis is that food images
have more informative regions than part-fixed fine-grained images.
However, it is challenging to capture so many informative regions
under the existing training mechanism of vanilla networks, since training
these regions together can not effectively learn each of them.

In this paper, we propose a self-boosting training mechanism, called
Progressive Self-Distillation (PSD), to learn more details for food
recognition. Instead of aggregating multiple detected local features,
the proposed PSD progressively enhances the ability of network to
mine other informative regions from an image with some informative
regions masked. To be more specific, the training of PSD simultaneously
contains multiple self-distillations. In each self-distillation, a
teacher network and a student network share an embedding network.
The student network receives a modified image from its teacher network
by removing some informative regions. Thus, the teacher network outputs
stronger semantic representations than the student network. Guided
by such teacher network with stronger semantics, the student network
is encouraged to mine other discriminative regions from modified images
by improving its own ability. Correspondingly, the ability of the
teacher network is also enhanced with the shared embedding network.
In the next self-distillation, more informative regions are masked
in the input image of the student network based on the previous self-distillation.
The multiple self-distillations are learned together by adjusting
their weights with a ramp-up function to achieve progressive training.
In this way, the teacher network incrementally improves its ability
to mine more discriminative regions. Only the teacher network is used
for inference without the help of the student network.

The proposed method can be flexibly implemented with different architectures
of the embedding network. Two representative architectures are employed,
including convolution neural networks (CNNs) and vision Transformers
(e.g., Swin Transformer \citep{liu2021swin}). Comprehensive experiments
on three benchmark datasets demonstrate the effectiveness and superiority
of our method.

The contributions of our paper are summarized as follows: 
\begin{itemize}
\item We propose a progressive self-distillation method, which progressively
mines more informative regions in a self-boosting manner for food
recognition. 
\item The proposed method is flexible to architectures of the embedding
network including convolution neural networks and vision Transformers.
\item We conduct extensive evaluation on three popular food benchmark datasets
to verify the effectiveness of the proposed method.
\end{itemize}

\section{Related Work}

\subsection{Food Recognition}

In the earlier years, some works use hand-crafted features \citep{lowe2004distinctive}
or combination of hand-crafted features \citep{martinel2015structured}
for food recognition. With the development of CNNs \citep{He-DRL-CVPR2016},
some works \citep{Kagaya-FDR-MM2014} directly employ various types
of CNNs for food recognition. 

Recently, some works \citep{Martinel-WSR-WACV2018,Qiu-MDFR-BMVC2019,min2020isia,Min-IGCMAN-ACMMM2019}
follow the key solution of fine-grained recognition, which firstly
locates informative regions in some weakly-supervised manners, and
then combines features of these regions with a global feature for
food recognition. For examples, the information of ingredient \citep{Min-IGCMAN-ACMMM2019}
is leveraged to extract local features, and these local features and
a global feature are combined to recognize food images. A slice network
\citep{Martinel-WSR-WACV2018} is proposed to capture specific vertical
food layers, and then combine the features of slice network with ones
from backbone network for food recognition. Although obtaining the
impressive performance, these methods possibly locate some incorrect
regions, which limits their effectiveness. Moreover, these methods
require to locate multiple regions and aggregate features of these
regions, resulting in high computational overhead in both training
phase and inference phase. Different from the above existing works
of aggregating multiple regions, the proposed PSD learns more informative
regions in a self-boosting manner. Compared with a standard classification
network, the PSD does not bring extra computational overhead in the
inference phase.

\subsection{Self-supervised Learning }

The approaches of self-supervised learning focus on designing auxiliary
objectives to learn useful feature representations by using the structure
of the data itself. The auxiliary objectives can be handcrafted pretext
tasks, such as relative patch prediction \citep{doersch2015unsupervised},
solving jigsaw puzzles \citep{noroozi2016unsupervised} and rotation
prediction \citep{komodakis2018unsupervised}. Although these methods
learn useful feature representations with big networks and long training
\citep{kolesnikov2019revisiting}, they heavily rely on somewhat adhoc
pretext tasks, which limits the generalization ability of learned
representations to some extent. 

Recently, contrastive learning \citep{hadsell2006dimensionality}
is increasingly successful in self-supervised learning. The core idea
of contrastive learning is to attract the positive sample pairs and
repulse the negative sample pairs. In practice, contrastive learning
methods benefit from a large number of negative samples, which can
be maintained in a memory bank \citep{wu2018unsupervised}. Without
requiring specialized architectures or a memory bank, a simple contrastive
self-supervised learning \citep{chen2020simple} is proposed while
this work requires a large batch size to work well. And a dynamic
dictionary \citep{he2020momentum} is used with a queue and a moving-averaged
encoder, building a large and consistent dictionary on-the-fly. To
eliminate the requirement of negative samples for reducing memory
consumption, a series of works \citep{chen2021exploring,grill2020bootstrap}
retain siamese architectures to learn invariant features by matching
positive samples, and employ a stop-gradient operation to prevent
model from collapsing. Interesting, the representations from contrastive
self-supervised pre-training can outperform their supervised counterparts
in certain tasks. In this paper, we borrow contrastive learning for
food recognition at semantic levels in a self-boosting manner to some
extent.

\subsection{Knowledge Distillation}

\begin{figure*}

\begin{centering}
\includegraphics[scale=0.675]{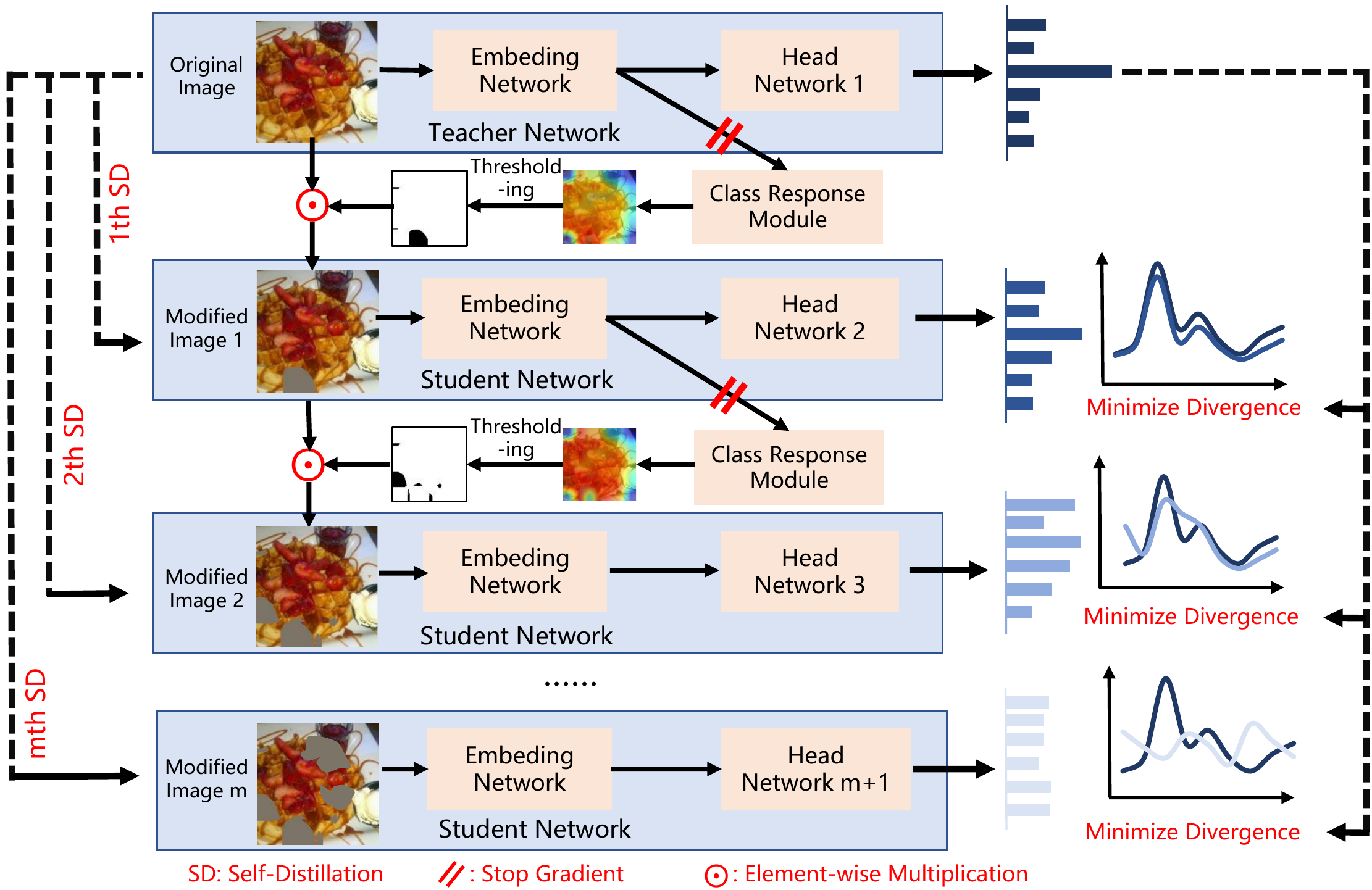}
\par\end{centering}

\caption{\label{framework_png}The training pipeline of the proposed method
PSD. The PSD contains multiple self-distillations, and in each self-distillation,
a teacher network and a student network share the same embedding network
but with different head networks.}
\end{figure*}

The concept of knowledge distillation was firstly proposed in \citep{hinton2015distilling}.
In a learning paradigm of knowledge distillation, a bigger teacher
network guides the training of a smaller student network to transfer
its knowledge to the student network. The knowledge can be distilled
via semantic distributions \citep{tung2019similarity,zhao2022decoupled,chen2022knowledge}
or intermediate features \citep{heo2019knowledge,tian2019contrastive}.
To enhance efficiency and effectiveness in knowledge transferring,
self knowledge distillation is proposed to utilize knowledge from
itself, without the involvement of extra networks. For example, aggregation
of various distortion data is used to achieve self-distillation \citep{lee2020self}.
Motivated by this, we propose a progressive self-distillation method
to mine more details for food recognition.

\section{The Proposed Method}

The proposed PSD progressively enhances the ability of network to
mine more discriminative regions for food recognition. As shown in
Figure \ref{framework_png}, the training pipeline of PSD includes
multiple self-distillations. In this section, we first present the
framework of self-distillation, and then introduce how to organize
multiple self-distillations for progressive training. Finally, the
method implementation is provided.

\subsection{Self-Distillation}

In self-distillation, there are a teacher network and a student network,
as shown in Figure \ref{framework_png}. The teacher network and the
student network share the same embedding network but with corresponding
head networks. In the teacher network, an original image is inputted.
An input image of a student network is modified from its teacher network
by removing some informative regions. Next, we will introduce how
to obtain an input image of a student network, and how to distill.

\textbf{Locating Discriminative Region.} The discriminative regions
are located with high class responses of feature map. Given an input
image $x$, an embedding network is used to extract a feature map
$f(x;\theta)=S\in\mathbf{R}{}^{H\times W\times D}$, where $f(;\theta)$
represents an embedding network with parameter $\theta$, $D$ is
the number of feature channels, and $H\times W$ is the spatial size
of the feature map. When using vision Transformer as an embedding
network, the tokens $T\times D$ are reshaped to a $H\times W\times D$
feature map. The feature map $S$ is then fed to a class response
module, which contains a global average pooling layer followed by
a fully connected layer. The weight matrix of the fully connected
layer is $\mathit{\Theta}\in\mathbf{R}{}^{D\times C}$, where $C$
is the number of food categories. Then a Class Response Map (CRM)
of the $c$-th category $M_{c}$ is computed as: 
\begin{equation}
\begin{aligned}M_{c}=\sum_{d=1}^{D}\mathit{\Theta}_{d,c}\times S_{d}\end{aligned}
\end{equation}
 where $\mathit{\Theta}_{d,c}$ represents the $d$-th weight for
the $c$-th category in $\mathit{\Theta}$, and $S_{d}$ is a $d$-th
feature map in $S$. The weight $\mathit{\Theta}$ needs to be learned,
and the optimization loss is:
\begin{equation}
L_{l}(x,y)=L_{ce}(\mathrm{GAP}(f(x;\theta))*\mathit{\Theta},y)\label{eq:crm}
\end{equation}
where $\mathrm{GAP()}$ is a global average pooling operation, $*$
is a matrix multiplication operation, $L_{ce}(,)$ is a cross-entropy
loss, and $y$ is the ground truth label. 
\begin{figure}
\noindent \begin{centering}
\includegraphics[scale=0.8]{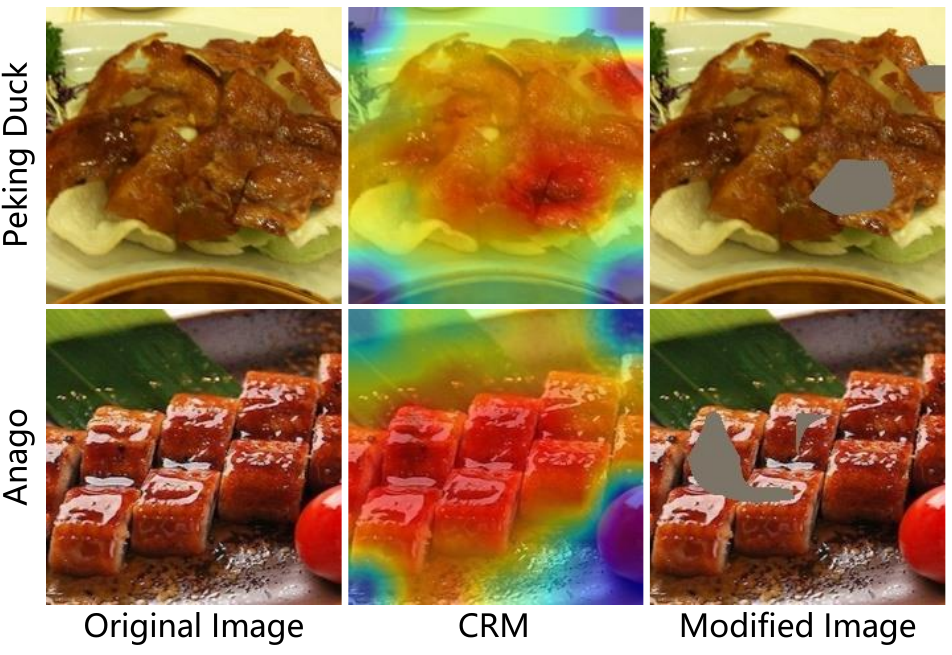}
\par\end{centering}
\caption{\label{fig:Some-examples-of mask images}Some examples of original
images, CRM and modified images.}
\end{figure}

The location of high value in the class response map $M_{c}$ can
represent discriminative area. Therefore, these high values in $M_{c}$
is used to locate discriminative regions. A certain percentile $\eta$
in $M_{c}$ is used as a threshold. Supposing the value of the percentile
is $\hat{w_{m}}$, the location of discriminative regions for the
$c$-th category is calculated as: 
\begin{equation}
\begin{aligned}Loc_{c}=\begin{cases}
1, & w_{m}\geq\hat{w_{m}}\\
0, & w_{m}<\hat{w_{m}}
\end{cases}\end{aligned}
\end{equation}
 where $w_{m}\in M_{c}$. 

\textbf{Masking Discriminative Region. }The ground truth label is
used to obtain the location of discriminative regions $Loc_{c}$,
and the discriminative regions are removed in image level. The formulation
of this process is defined as follows: 
\begin{equation}
\bar{x}=(1-Loc_{c})\odot x,
\end{equation}
where $\odot$ means an element-wise multiplication, and $\bar{x}$
is a modified image. As shown in Figure \ref{fig:Some-examples-of mask images},
the discriminative regions are masked to form modified images.

\textbf{Distillation Objective.} The image $x$ is fed into a teacher
network, and the modified image $\bar{x}$ is fed into a student network.
Therefore, the teacher network outputs stronger semantic information
than the student network. The optimization goal is to minimize difference
in output distribution between the teacher network and the student
network, and the loss is as follows:
\begin{equation}
L_{d}(x,\bar{x})=D_{KL}(h(f(x;\theta);\phi_{t}),h(f(\bar{x};\theta);\phi_{s})))\label{eq:kl}
\end{equation}
where $D_{KL}(,)$ is Kullback-Leibler divergence, $h(;)$ represents
a head network, and $\phi_{t}$ and $\phi_{s}$ are parameters of
$h(;)$ in the teacher network and the student network, respectively.

Guided by stronger semantic information, the student network is encouraged
to mine other discriminative regions from the modified image $\bar{x}$,
improving the ability of the student network. The ability of the teacher
network is also enhanced since the teacher network and the student
network share the same embedding network.

\subsection{Progressive Training}

Progressive training methodology has been widely utilized for image
generation tasks \citep{ahn2018image,karras2019style}. This methodology
starts with generating low-resolution images, and then gradually increases
the generated resolution. In this methodology, an easy task is completed
first, and then the difficulty of the task is gradually increased.
Our work uses a progressive training methodology to organize multiple
self-distillations. In these self-distillations, an input image of
the student network is a modified image of student network in the
previous self-distillation, except for the first time. In the first
self-distillation, the input image of student network is a modified
image of the teacher network.

The classification loss is:
\begin{equation}
\begin{aligned}L_{g}(x,y)=L_{ce}(h(f(x;\theta);\phi_{t}),y)\end{aligned}
\end{equation}
Combining with the classification loss, the locating loss $L_{l}$
in Eq. \ref{eq:crm} and the distillation loss $L_{d}$ in Eq. \ref{eq:kl},
the final optimization loss is:
\begin{equation}
\begin{aligned}L=L_{g}(x,y)+\mathop{\sum_{i=1}^{m}(\omega_{l}}L_{l}(\bar{x}_{i-1},y)+\omega_{d}L_{d}(x,\bar{x}_{i}))\end{aligned}
\label{eq:total loss}
\end{equation}
where $\bar{x}_{i}$ is an input image of the student network in $i$-th
self-distillation, $\bar{x}_{0}$ is $x$, $m$ is the number of self-distillation,
and $\omega_{l}$ and $\omega_{d}$ are balance parameters. Instead
of using a step-by-step progressive training like \citep{ahn2018image,karras2019style},
multiple self-distillations are learned together by adjusting the
value of $\omega_{d}$ during training phase. The $\omega_{d}$ starts
from a small value to a fixed value $\alpha$ with a ramp-up function
\citep{laine2016temporal}, and it can be formalized as follows:
\begin{equation}
\begin{aligned}\omega_{d}=\begin{cases}
\alpha\ast exp(-5(1-\frac{e}{\beta})^{2}), & e<\beta\\
\alpha, & e\geq\beta
\end{cases}\end{aligned}
\label{eq:wd}
\end{equation}
where $e$ denotes the current epoch during training phase, $\alpha$
is a scalar, and $\beta$ is an integer. 

At the begin of training, the predictions of the teacher network may
be incorrect. A small value of $\omega_{d}$ can prevent the student
network learning incorrect knowledge from the teacher network. After
the predictions of the teacher network become confident, a big value
of $\omega_{d}$ can encourage the student network to learn correct
knowledge from the teacher network. Using a ramp-up $\omega_{d}$
makes multiple self-distillations of different difficulties learned
together. By using this progressive training, the teacher network
incrementally improves the ability to mine more discriminative regions
for food recognition. 

\subsection{Implementation}

The embedding network can be implemented with the vast majority of
networks. In this work, we use two types of network architectures
including vision transformers and CNN-based networks. The head network
contains a global average pooling layer and a fully connected layer.
The class response module acts on the feature map of the third hidden
layer in the embedding network. A stop-gradient operation is performed
on the feature map before inputting the class response module. Therefore,
the loss in class response module (i.e., Eq. \ref{eq:crm}) does not
affect the training of the embedding network. Although our method
PSD requires multi-stage network calculation in training phase, only
one stage network calculation (i.e., the teacher network) is used
for classification in inference phase. Compared with a standard classification
network, the PSD does not bring additional computational overhead
in the inference phase.

\section{Experiments}

We will validate the effectiveness of the proposed method by answering
the following two questions: Q1. Is the proposed method PSD effective
for food recognition? Q2. Does the proposed method PSD learn more
discriminative regions? In this section, the experimental setup is
firstly introduced. Next, the two questions are answered in experimental
results and analysis. Finally, the further analysis is provided.

\subsection{Experimental Setup}

\textbf{Datasets.} We validate our method on three commonly used food
datasets. 
\begin{itemize}
\item \textbf{ETHZ Food-101} \citep{Bossard-Food101-ECCV2014} contains
101,000 images with 101 food categories. Each category has 1,000 images
including 750 training images and 250 test images. 
\item \textbf{Vireo Food-172 }\citep{Chen-DIRCRR-MM2016} contains 110,241
food images from 172 categories. Following commonly used splits, 60\%,
10\%, 30\% images of each food category are randomly selected for
training, validation and testing, respectively. 
\item \textbf{ISIA Food-500} \citep{min2020isia} consists of 399,726 images
with 500 categories. The average number of images per category is
about 800. This dataset is divided into 60\%, 10\% and 30\% images
for training, validation and testing, respectively. 
\end{itemize}
\textbf{Implementation Details. }All experiments are implemented on
the Pytorch platform with one Nvidia A100 GPU. The input image size
is set to $224\times224$ in all experiments. For fair comparisons
and re-implementations, the same random seed is used to eliminate
the training bias in all experiments. Top1 accuracy and Top5 accuracy
are used as evaluation metrics. On Vireo Food-172 and ISIA Food-500,
the model of the highest performance on validation set is used for
test. On ETHZ Food-101, since there is no validation set, the last
model is used for test. We set a percentile $\eta=5\%$ in $M_{c}$
as a threshold, $\omega_{l}=1$, the ramp-up epochs $\beta=5$ in
Eq. \ref{eq:total loss}, and the number of self-distillation $m=2$
in all experiments.

\begin{table}

\noindent \centering{}%

\begin{tabular}{ll|cc}
\hline 
\multicolumn{2}{l|}{Method} & Top1 & Top5\tabularnewline
\hline 
\multirow{6}{*}{MF} & \small{WISeR \citep{Martinel-WSR-WACV2018}$^{\dagger}$} & 90.27 & 98.71\tabularnewline
 & IG-CMAN{\small{} \citep{Min-IGCMAN-ACMMM2019}} & 90.37 & 98.42\tabularnewline
 & PAR-Net{\small{} \citep{Qiu-MDFR-BMVC2019}}$^{\dagger}$ & 90.40 & -\tabularnewline
 & MSMVFA{\small{} \citep{Min-MSMVFA-TIP2019}} & 90.59 & 98.25\tabularnewline
 & SGLANet{\small{} \citep{min2020isia}}$^{\dagger}$ & 90.92 & 98.24\tabularnewline
 & IGRL{\small{} \citep{Wang2022}} & 92.36 & 98.68\tabularnewline
\hline 
\multirow{12}{*}{SF} & SFLR{\small{} \citep{Bola2017Simultaneous}} & 79.20 & 94.11\tabularnewline
 & SENet154 {\small{}\citep{Jie2017Squeeze}} & 88.62 & 97.57\tabularnewline
 & DLA{\small{} \citep{Yu-DLA-CVPR2018}}$^{\ast}$ & 90.00 & -\tabularnewline
 & Incep-Res-v2{\small{} \citep{Cui-LSFGC-CVPR2018}} & 90.40 & -\tabularnewline
 & Incep-v4{\small{} \citeyearpar[Kornblith et al.][]{Kornblith2018Do}}$^{\ast}$ & 90.80 & -\tabularnewline
 & GPipe{\small{} \citep{huang2019gpipe}} & 93.00 & -\tabularnewline
 & EfficientNet {\small{}\citep{tan2019efficientnet}} & 93.00 & -\tabularnewline
 & Grafit {\small{}\citep{touvron2021grafit}} & 93.70 & -\tabularnewline
\cline{2-4} \cline{3-4} \cline{4-4} 
 & DenseNet161{\small{} \citep{huang2017densely}} & 86.93 & 97.10\tabularnewline
 & DenseNet161+PSD (our) & 87.40 & 97.20\tabularnewline
\cline{2-4} \cline{3-4} \cline{4-4} 
 & Swin-B {\small{}\citep{liu2021swin}} & 93.91 & 99.03\tabularnewline
 & Swin-B+PSD (our) & \textbf{94.56} & \textbf{99.34}\tabularnewline
\hline 
\end{tabular}

\caption{\label{tab:performance_food101}Accuracy (\%) comparison on ETHZ Food-101.
$^{\dagger}$: 10 crop images for test. $^{\ast}$: $448\times448$
resolution images. MF: multi-feature aggregation. SF: single feature
for test.}
\end{table}

When employing Swin-B \citep{liu2021swin} as an embedding network,
the model is optimized by adamw \citep{kingma2014adam} algorithm
with an initial learning rate of $5\times10^{-5}$ and a weight decay
of $10^{-8}$. The model is initialized with ImageNet-22K pre-trained
model. A cosine decay learning rate scheduler with 5 epochs of warm-up
is used. The total number of training epochs is 50, and a batch size
of 42 and gradient clipping with a max norm of 5 are used. Similar
to the work \citep{liu2021swin}, gradient accumulation step of 2
is used to reduce GPU consumption and stochastic depth ratio of 0.2
is adopted. In Eq. \ref{eq:wd}, $\alpha=2.0$.

When employing DenseNet161 \citep{huang2017densely} as an embedding
network, the model is optimized using stochastic gradient descent
with a momentum of 0.9 and a weight decay of $10^{-4}$. The model
is initialized with ImageNet-1K pre-trained model. The learning rate
is initially set to $10^{-3}$ and divided by 10 after 10 epochs.
The total number of training epochs is 30, and the batch size is 42.
In Eq. \ref{eq:wd}, $\alpha=1.0$.

\subsection{Experimental Results and Aanlysis}

\textbf{Evaluation on Three Benchmark Datasets.} The experimental
results on ETHZ Food-101 are illustrated on Table \ref{tab:performance_food101}.
We build a strong baseline with Swin Transformer \citep{liu2021swin}.
The baseline method Swin-B has already outperformed some previous
best methods, including some methods of multi-feature aggregation
(e.g., SGLANet \citep{min2020isia} and PAR-Net \citep{Qiu-MDFR-BMVC2019})
and some methods of using advanced architectures (e.g., Grafit \citep{touvron2021grafit}
and GPipe \citep{huang2019gpipe}). Although a strong baseline
is achieved with the Swin-B, the proposed PSD still obtains 0.65\%
improvements in Top1 accuracy and 0.31\% improvements in Top5 accuracy,
exhibiting its superiority. When using DenseNet161, the proposed PSD
obtains about 0.5\% gains in Top1 accuracy. 

\begin{table}

\noindent \centering{}%
\begin{tabular}{ll|cc}
\hline 
\multicolumn{2}{l|}{Method} & Top1 & Top5\tabularnewline
\hline 
\multirow{6}{*}{MF} & IG-CMAN{\small{} \citep{Min-IGCMAN-ACMMM2019}} & 90.63 & 98.40\tabularnewline
 & PAR-Net{\small{} \citep{Qiu-MDFR-BMVC2019}}$^{\dagger}$ & 90.20 & -\tabularnewline
 & MSMVFA{\small{} \citep{Min-MSMVFA-TIP2019}} & 90.61 & 98.31\tabularnewline
 & AFN{\small{} \citep{liu2020food}} & 89.54 & 98.05\tabularnewline
 & SGLANet{\small{} \citep{min2020isia}}$^{\dagger}$ & 90.98 & 98.35\tabularnewline
 & MVANET{\small{} \citep{liang2021mvanet}} & 91.08 & 98.86\tabularnewline
\hline 
\multirow{7}{*}{SF} & VGG16 {\small{}\citep{Szegedy-GDC-CVPR2015}} & 80.41 & 95.59\tabularnewline
 & MTDCNN{\small{} \citep{Chen-DIRCRR-MM2016}} & 82.06 & 95.88\tabularnewline
 & SENet154{\small{} \citep{Jie2017Squeeze}} & 88.71 & 97.74\tabularnewline
\cline{2-4} \cline{3-4} \cline{4-4} 
 & DenseNet161{\small{} \citep{huang2017densely}} & 88.25 & 97.53\tabularnewline
 & DenseNet161+PSD (our) & 89.00 & 97.70\tabularnewline
\cline{2-4} \cline{3-4} \cline{4-4} 
 & Swin-B{\small{} \citep{liu2021swin}} & 92.75 & 98.71\tabularnewline
 & Swin-B+PSD (our) & \textbf{92.91} & \textbf{99.08}\tabularnewline
\hline 
\end{tabular}

\caption{\label{tab:performance_food172}Accuracy (\%) comparison on Vireo
Food-172. $^{\dagger}$: 10 crop images for test.}
\end{table}

The experimental results on Vireo Food-172 are illustrated on Table
\ref{tab:performance_food172}. A strong baseline is built with Swin-B,
which outperforms the best method MVANET \citep{liang2021mvanet}
by 1.67\% in Top1 accuracy, and is comparable with MVANET in Top5
accuracy. Based on this baseline, the proposed Swin-B+PSD still obtains
performance improvements. When using DenseNet161, the proposed PSD
achieves 0.75\% gains in Top1 accuracy.

The experimental results on ISIA Food-500 are illustrated on Table
\ref{tab:performance_food500}. A strong baseline is also built with
Swin-B, which outperforms the best method TPSKG \citep{liu2022transformer} by about 1.9\% in Top1 accuracy and SGLANet \citep{min2020isia}
by over 1.0\% in Top5 accuracy. Note that TPSKG also employs vision
Transformer as backbone and use higher resolution images. Despite
achieving a strong baseline, the proposed method still obtains relatively
large performance gains, i.e., 2.78\% improvements in Top1 accuracy
and 2.57\% improvements in Top5 accuracy, validating the advantage
of the proposed method. When using DenseNet161, the proposed PSD achieves
about 0.9\% gains in Top1 accuracy and 1.24\% gains in Top5 accuracy.

\begin{table}

\noindent \centering{}%
\begin{tabular}{ll|cc}
\hline 
\multicolumn{2}{l|}{Method} & Top1 & Top5\tabularnewline
\hline 
\multirow{3}{*}{MF} & NTS-NET{\small{} \citep{Yang-L2Nav-ECCV2018}} & 63.66 & 88.48\tabularnewline
 & WS-DAN{\small{} \citep{Hu_2019_CVPR}} & 60.67 & 86.48\tabularnewline
 & SGLANet{\small{} \citep{min2020isia}} & 64.74 & 89.12\tabularnewline
\hline 
\multirow{11}{*}{SF} & VGG16{\small{} \citep{Szegedy-GDC-CVPR2015}} & 55.22 & 82.77\tabularnewline
 & GoogLeNet{\small{} \citep{Meyers-Im2Calories-ICCV2015}} & 56.03 & 83.42\tabularnewline
 & ResNet152{\small{} \citep{He-DRL-CVPR2016}} & 57.03 & 83.80\tabularnewline
 & WRN50{\small{} \citep{Zagoruyko-WRN-BMVC2016}} & 60.08 & 85.98\tabularnewline
 & SENet154{\small{} \citep{Jie2017Squeeze}} & 63.83 & 88.61\tabularnewline
 & DCL{\small{} \citep{Chen_2019_CVPR}} & 64.10 & 88.77\tabularnewline
 & TPSKG{\small{} \citep{liu2022transformer}}$^{\ast}$ & 65.40 & -\tabularnewline
\cline{2-4} \cline{3-4} \cline{4-4} 
 & DenseNet161{\small{} \citep{huang2017densely}} & 60.05 & 86.09\tabularnewline
 & DenseNet161+PSD (our) & 60.94 & 87.33\tabularnewline
\cline{2-4} \cline{3-4} \cline{4-4} 
 & Swin-B{\small{} \citep{liu2021swin}} & 67.32 & 90.18\tabularnewline
 & Swin-B+PSD (our) & \textbf{70.10} & \textbf{92.75}\tabularnewline
\hline 
\end{tabular}

\caption{\label{tab:performance_food500}Accuracy (\%) comparison on ISIA Food-500.
$^{\ast}$: $448\times448$ resolution images.}
\end{table}

To summarize, the proposed method PSD obtains performance gains on
the above three datasets (e.g., ETHZ Food-101, Vireo Food-172 and
ISIA Food-500) in two architectures (e.g., Swin-B and DenseNet161).
This can validate that the proposed method is effective for food recognition
(Q1). 

\begin{figure}
\begin{centering}
\includegraphics[scale=0.61]{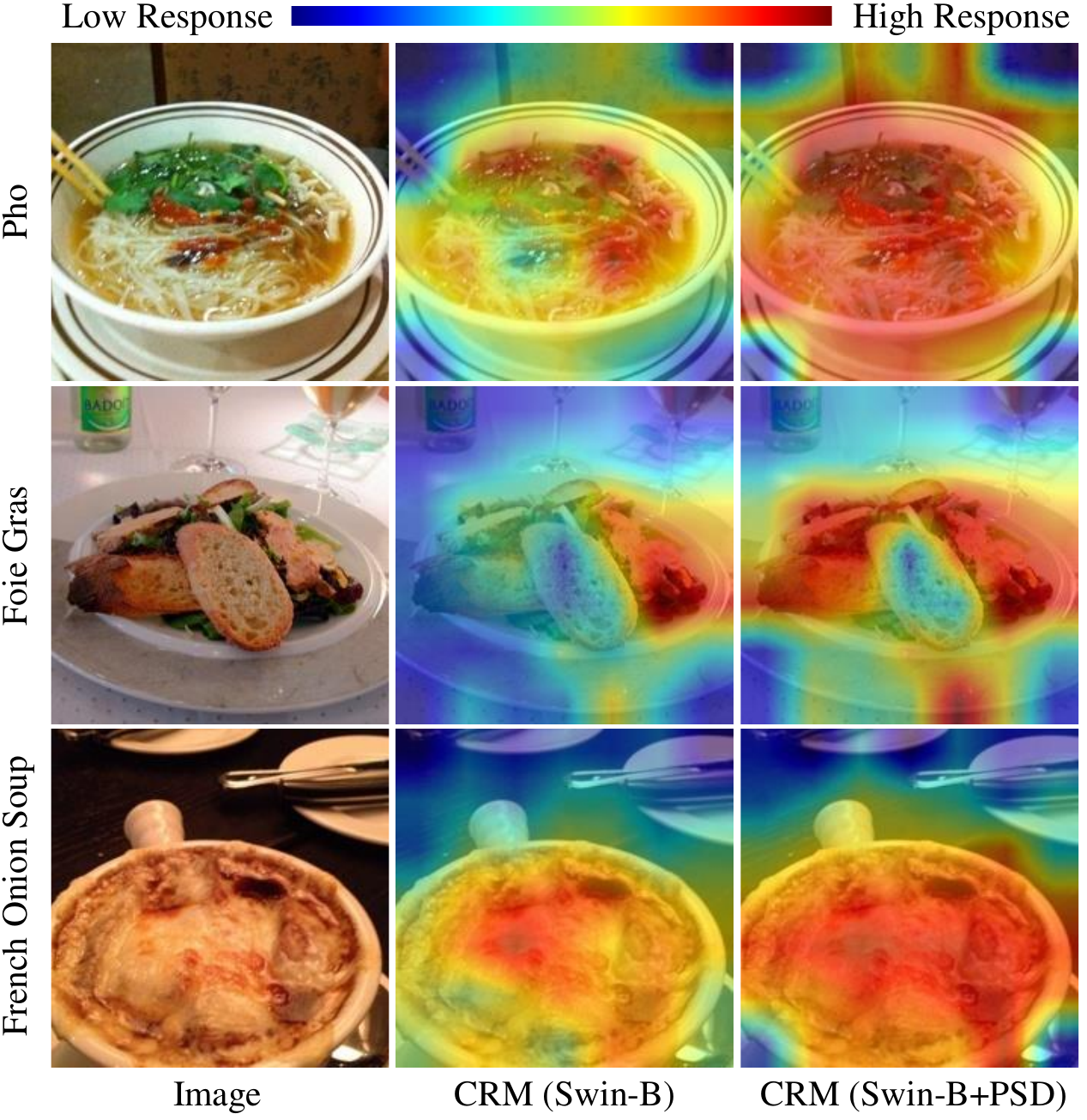}
\par\end{centering}
\caption{\label{fig:crm_examples} Some visualizations of Class Response Map.}
\end{figure}

\textbf{Visualization of Response Map}. To demonstrate more informative
regions learned by the proposed PSD, we visualize the class response
map of both Swin-B and Swin-B+PSD, as shown in Figure \ref{fig:crm_examples}.
We can obviously observe that the proposed Swin-B+PSD obtains more
high-response regions than Swin-B. These high-response regions are
also discriminative, which can explain that the proposed method learns
more discriminative regions (Q2).

\begin{figure}
\noindent \begin{centering}
\includegraphics[scale=0.62]{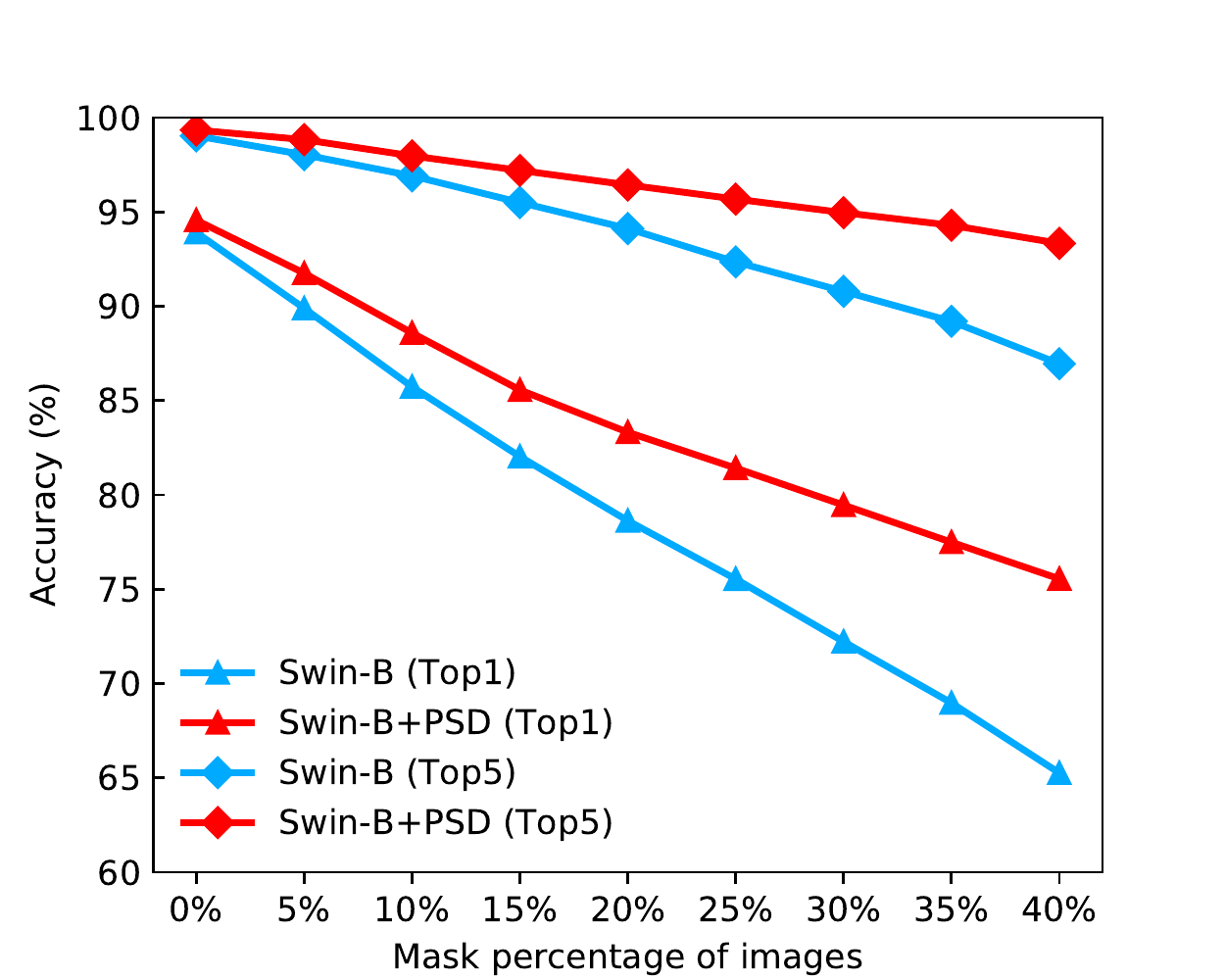}
\par\end{centering}
\caption{\label{fig:mask_results}Accuracy of masked images on ETHZ
Food-101.}
\end{figure}

\textbf{Test on Masked Images.} Furthermore, we test the trained model
with modified images where some regions are randomly masked. The mask
percentage for each test image ranges from 0\% to 40\%. The experimental
results on ETHZ Food-101 are exhibited in Figure \ref{fig:mask_results}.
Both the baseline method Swin-B and the proposed Swin-B+PSD degrade
performance with increasing mask percentage of each test image. Top1
and Top5 accuracy of Swin-B drops over 25\% and 10\%, respectively.
In contrast, Top1 and Top5 accuracy of Swin-B+PSD drops less than
20\%, and 7\%, respectively. The performance drop of the proposed
PSD is lower than the baseline method in both Top1 accuracy and Top5
accuracy. This can explain that the proposed PSD learns more effective
features than the baseline method from masked images. To some extent,
this also can explain that the proposed PSD learns more discriminative
regions (Q2).

\subsection{Further Analysis }

\textbf{Comparisons with Numbers of Self-distillations.} The experimental
results of different numbers of self-distillations on ETHZ Food-101
and ISIA Food-500 are illustrated in Table \ref{tab:Accuracy-different SD}.
Compared with the baseline method Swin-B, the proposed method equipped
with 1, 2, or 3 self-distillations obtains improvements, demonstrating
the effectiveness of the proposed method. Among them, the proposed
method equipped with 2 self-distillations achieves the best performance.
Based on two self-distillations, adding a third self-distillation
results in performance drops. The third self-distillation is a tougher
task than the first two, since it removes more informative regions.
This tough task possibly increases the difficulty of training. 

\begin{table}

\begin{spacing}{0.9}
\renewcommand{\arraystretch}{1.1}
\noindent \centering{}%
\begin{tabular}{l|c|c|c|c}
\hline 
\multirow{2}{*}{Method} & \multicolumn{2}{c|}{Food-101 (\%)} & \multicolumn{2}{c}{Food-500 (\%)}\tabularnewline
 & Top1 & Top5 & Top1 & Top5\tabularnewline
\hline 
Swin-B & 93.91 & 99.03 & 67.32 & 90.18\tabularnewline
Swin-B+PSD {\small{}($m$=1)} & 94.46 & 99.25 & 69.47 & 92.24\tabularnewline
Swin-B+PSD {\small{}($m$=2)} & 94.56 & 99.34 & 70.10 & 92.75\tabularnewline
Swin-B+PSD {\small{}($m$=3)} & 94.44 & 99.30 & 69.94 & 92.63\tabularnewline
\hline 
\end{tabular}
\end{spacing}
\caption{\label{tab:Accuracy-different SD}Comparisons with numbers of self-distillations.}
\end{table}

\textbf{Comparisons with Settings of Progressive Training.} We consider
the following two variants of progressive training, including i) step-by-step
progressive training: multiple self-distillations are trained from
first to last, ii) head shared progressive training: the teacher network
and the student network share the head network. The experimental results
of two variants are illustrated at the top of Table \ref{tab:Comparison-with-different types}.
Our progressive training obtains better performance than the two variants,
demonstrating its advantages.

\begin{table}

\begin{spacing}{0.9}
\renewcommand{\arraystretch}{1.1}
\noindent \centering{}%
\begin{tabular}{l|c|c|c|c}
\hline 
\multirow{2}{*}{Method} & \multicolumn{2}{c|}{Food-101 (\%)} & \multicolumn{2}{c}{Food-500 (\%)}\tabularnewline
 & Top1 & Top5 & Top1 & Top5\tabularnewline
\hline
Swin-B+PSD {\small{}(SbS)} & 94.40 & 99.30 & 69.30 & 92.01\tabularnewline
Swin-B+PSD {\small{}(HS)} & 94.39 & 99.28 & 69.79 & 92.47\tabularnewline
Swin-B+PSD {\small{}(our)} & 94.56 & 99.34 & 70.10 & 92.75\tabularnewline
\hline 
Swin-B & 93.91 & 99.03 & 67.32 & 90.18\tabularnewline
Swin-B {\small{}(DA Mask5\%)} & 93.87 & 98.90 & 67.41 & 90.13\tabularnewline
Swin-B {\small{}(DA Mask10\%)} & 93.88 & 98.94 & 67.43 & 90.12\tabularnewline
\hline 
\end{tabular}
\end{spacing}

\caption{\label{tab:Comparison-with-different types}Comparisons with progressive
training and Data Augmentation (DA). SbS: step-by-step. HS:head shared.}
\end{table}

\textbf{Comparisons with Data Augmentation.} Masking discriminative
regions from images with the class response module is considered as
a manner of data augmentation. The experimental results of this data
augmentation are illustrated at the bottom of Table \ref{tab:Comparison-with-different types}.
This data augmentation does not bring performance improvements based
on Swin-B. This can illustrate that the performance improvement obtained
by our method PSD is not due to the use of modified images.

\textbf{Comparisons with Other Learning Methods.} We compare the proposed
method with other learning methods BAN \citep{furlanello2018born}
and MutL \citep{zhang2018deep}. The two works use the same backbone (i.e., Swin-B), batch size and the settings of optimizer as ours. Experimental results are shown in Table
\ref{tab:Comparisons-with-other}. The proposed method obtains higher accuracy
than the two works on ETHZ Food-101 and ISIA Food-500, illustrating its superiority. 
\begin{table}

\begin{spacing}{0.9}
\renewcommand{\arraystretch}{1.1}
\noindent \centering{}%
\begin{tabular}{l|c|c|c|c}
\hline
\multirow{2}{*}{Method} & \multicolumn{2}{c|}{Food-101 (\%)} & \multicolumn{2}{c}{Food-500 (\%)}\tabularnewline
 & Top1 & Top5 & Top1 & Top5\tabularnewline
\hline 
BAN & 94.10 & 99.02 & 68.01 & 91.15\tabularnewline
MutL & 94.29 & 99.23 & 69.23 & 92.22\tabularnewline
Our & 94.56 & 99.34 & 70.10 & 92.75\tabularnewline
\hline 
\end{tabular}
\end{spacing}
\begin{spacing}{0.3}
\caption{\label{tab:Comparisons-with-other}Comparsions with other learning
methods.}
\end{spacing}

\end{table}

\begin{figure}
\begin{centering}
\includegraphics[scale=0.58]{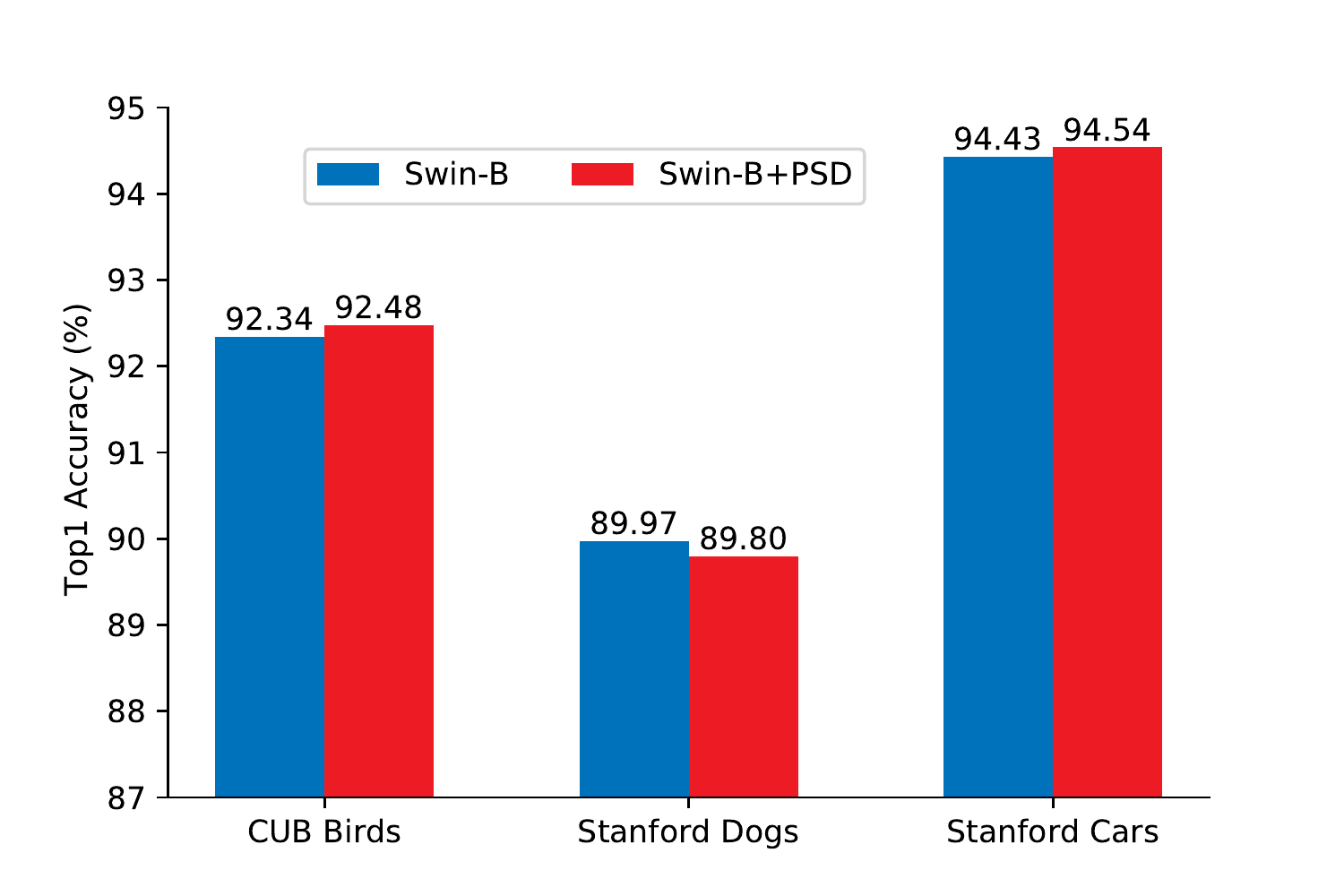}
\par\end{centering}
\caption{\label{fig:Top1-accuracy-cub}Top1 accuracy comparison on CUB Birds,
Stanford Dogs, and Stanford Cars.}

\end{figure}

\textbf{Experiments on Part-fixed Fine-grained Recognition.} We employ
the proposed method on three commonly used fine-grained datasets (i.e.,
CUB Birds \citep{wah2011caltech}, Stanford Dogs \citep{KhoslaYaoJayadevaprakashFeiFei_FGVC2011}
and Stanford Cars \citep{krause20133d}). The experimental results
are shown in Figure \ref{fig:Top1-accuracy-cub}. Based on Swin-B,
our method hardly obtains performance gains on three part-fixed fine-grained
datasets. Compared with food images, the number of discriminative
regions in part-fixed fine-grained images is limited, making it difficult
to mine more discriminative details.

\section{Conclusions}

In this paper, we propose a Progressive Self-Distillation method (PSD)
to learn more details for food recognition. Instead of locating multiple
regions, the proposed PSD progressively enhances the ability of network
to mine more informative regions. The training of PSD simultaneously
contains multiple self-distillations. In each self-distillation, the
student network is guided with stronger semantic representations from
the teacher network to improve its ability. The ability of the teacher
network also be improved with a shared embedding network in the student
network. By using progressive training to organize multiple self-distillations,
the teacher network incrementally improves the ability to mine more
discriminative regions. Comprehensive experiments on three benchmark
datasets demonstrate the effectiveness of our proposed method.

\section{Acknowledgments}
 This work was supported in part by the National Natural Science Foundation of China under Grant 62202058.
\bibliography{aaai23}

\end{document}